\documentclass[10pt,twocolumn,letterpaper]{article}

\usepackage{cvpr}
\usepackage{times}
\usepackage{epsfig}
\usepackage{graphicx}
\usepackage{verbatim}
\usepackage{url}            
\usepackage{booktabs}       
\usepackage{amsfonts}       
\usepackage{nicefrac}       
\usepackage{microtype}      
\usepackage{amssymb,amsmath,amsthm}
\usepackage{subcaption}
\usepackage{fullpage}
\usepackage{floatrow}
\usepackage{tabu}
\newcommand{\argmax}{\operatornamewithlimits{argmax}}
\usepackage{enumitem}
\usepackage{multirow}
\usepackage{bm}

\usepackage[pagebackref=true,breaklinks=true,letterpaper=true,colorlinks,bookmarks=false]{hyperref}

\cvprfinalcopy 

\def\cvprPaperID{} 

\ifcvprfinal\fi

\title{From source to target and back: Symmetric Bi-Directional Adaptive GAN}

\author{
  Paolo Russo$^1$, Fabio M. Carlucci$^1$, Tatiana Tommasi$^2$ and Barbara Caputo$^{1,2}$\\
  $^1$Department DIAG, Sapienza University of Rome, Italy \\
  $^2$Italian Institute of Technology\\
  {\tt\small \{prusso,fabiom.carlucci,caputo\}@dis.uniroma1.it, tatiana.tommasi@iit.it}
}
\begin{document}
\maketitle

\begin{abstract}
The effectiveness of GANs in producing images according to a specific 
visual domain has shown potential in  unsupervised domain adaptation. 
Source labeled images have been modified to mimic target samples for 
training classifiers in the target domain, 
and inverse mappings from the target to the source domain have also been evaluated, 
without new image generation. 

In this paper we aim at getting the best of both worlds
by introducing a symmetric mapping among domains. We jointly optimize bi-directional 
image transformations combining them with target self-labeling. 
We define a new class consistency loss that aligns the generators in the two directions, 
imposing to preserve the class identity of an image passing through both domain mappings.
A detailed analysis of the reconstructed images, a thorough ablation study
and extensive experiments on six different settings confirm the power
of our approach. 
\end{abstract}

\section{Introduction}
\label{sec:intro}
The ability to generalize across domains is challenging when there is ample labeled data 
on which to train a deep network (source domain), 
but no annotated data for the target domain. 
To attack this issue, 
a wide array of methods have been proposed, 
most of them aiming at reducing the shift between the source and target distributions (see Sec. \ref{sec:related} for a review of previous work). An alternative is
mapping the source data into the target domain,
either by modifying the image representation 
\cite{Ganin:DANN:JMLR16} or by directly generating a new version of the source images \cite{Bousmalis:Google:CVPR17}. 
Several authors proposed approaches that follow both these strategies by building over 
Generative Adversarial Networks (GANs) \cite{Goodfellow:GAN:NIPS2014}. 
A similar but inverse method maps the target data into the source domain, where there is already an abundance of labeled images \cite{Hoffman:Adda:CVPR17}.

We argue that these two mapping directions should not be 
alternative, but 
complementary. 
Indeed, the main ingredient for adaptation is the ability of transferring successfully the style of 
one domain to the images of the other. This, given a fixed generative architecture, will 
depend on the application: there may be cases where mapping from the source to the target is 
easier, and cases where it is true otherwise. By pursuing both directions in a 
unified architecture, we can obtain a system more robust and more general than previous adaptation algorithms. 

\begin{figure*}
\includegraphics[width=0.98\textwidth]{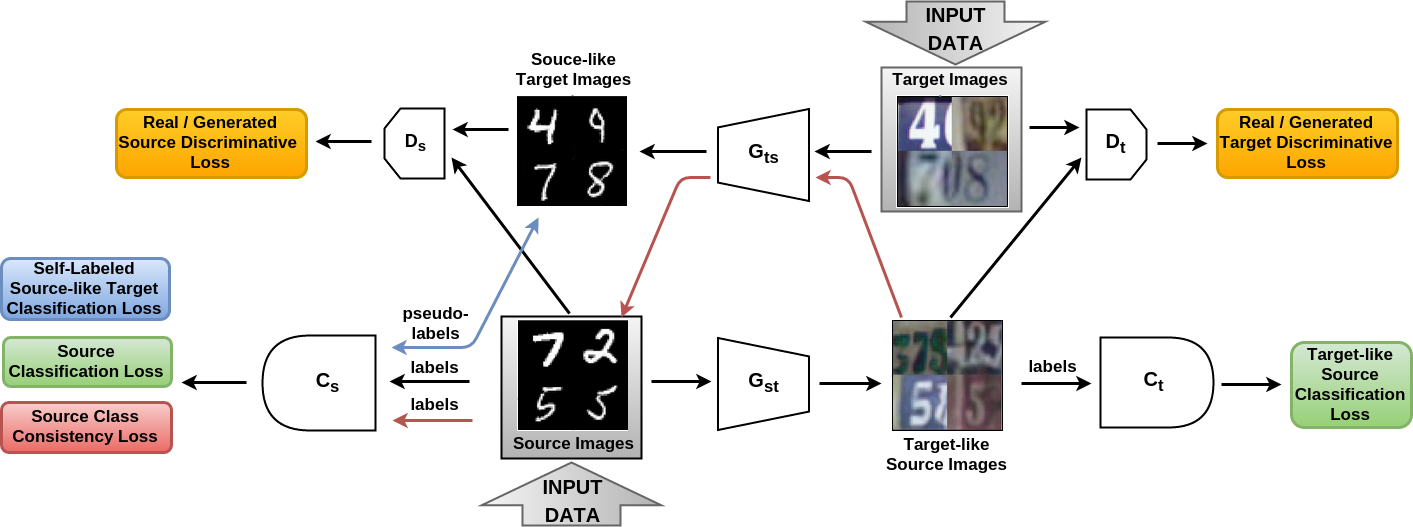}
{\caption{SBADA-GAN, training: the data flow starts from the source and target images
indicated by the Input Data arrows. The bottom and top row 
show respectively the source-to-target and target-to-source symmetric directions. The generative models
 $G_{st}$ and $G_{ts}$ transform the source images to the target domain and  vice versa.
 $D_{s}$ and $D_{t}$ discriminate real from generated images of source and target.
 Finally the classifiers  $C_{s}$ and $C_{t}$ are trained to recognize respectively the original source images  
 and their target-like transformed versions.
The bi-directional blue arrow indicates that the 
source-like target images are automatically annotated and the assigned pseudo-labels
are re-used by the classifier $C_s$. The red arrows describe the class consistency
condition by which source images transformed to the target domain through $G_{st}$ and back to the source domain
through $G_{ts}$ should maintain their ground truth label.}

\label{fig:sbadagan}\vspace{-1mm}}
\end{figure*}

With this idea in mind, we designed SBADA-GAN: Symmetric Bi-Directional ADAptive Generative Adversarial Network.
Its distinctive features are 
(see Figure \ref{fig:sbadagan} for a schematic overview):
\begin{itemize}[leftmargin=*]
\vspace{-1mm}
\item it exploits two generative adversarial losses that encourage the network to produce 
target-like images from the source samples and source-like images from the target samples.
Moreover, it jointly minimizes two classification losses, one on the original 
source images and the other on the transformed target-like source images;
\vspace{-3mm}
\item it uses the source classifier to annotate the source-like transformed target images. 
Such pseudo-labels 
help regularizing the same classifier while improving 
the target-to-source generator model by backpropagation;
\vspace{-1mm}
\item it introduces a new semantic constraint on the source images: the \textit{class consistency loss}. It 
imposes that by mapping source images towards the target domain and then again towards the 
source domain they should get back to their ground truth class.
This last condition is less restrictive than a standard reconstruction loss 
\cite{CycleGAN2017,DBLP:conf/icml/KimCKLK17}, as it deals 
only with the image annotation and not with the image appearance. Still, our experiments 
show that it is highly effective in aligning the domain mappings in the two directions;  
\vspace{-1mm}
\item at test time the two trained classifiers are used respectively
on the original target images and on their source-like transformed version. The two 
predictions are integrated to produce the final annotation.
\end{itemize}

Our architecture yields realistic image reconstructions while  competing against previous 
state-of-the-art classifiers and exceeding them on four out of six different 
unsupervised adaptation settings. 
An ablation study showcasing the importance of each component in the architecture, and investigating 
the robustness with respect to its hyperparameters,
sheds light on the inner workings of the approach, while providing  further evidence of its value.

\section{Related Work}
\label{sec:related}
\paragraph{GANs}
Generative Adversarial Networks are composed of two modules, 
a generator and a discriminator. The generator's objective is to synthesize samples whose 
distribution closely matches that of real data, while the discriminator objective is to 
distinguish real from generated samples. 
GANs are agnostic 
to the training samples labels, while conditional 
GAN variants \cite{Mirza:cGAN:arXiv2014} exploit the class annotation as additional information 
to both the generator and the discriminator.
Some works 
used multiple GANs: in CoGAN \cite{Liu:coGAN:NIPS16} two generators 
and two discriminators are coupled by weight-sharing to learn the joint distribution of images in two 
different domains without using pair-wise data.
Cycle-GAN \cite{CycleGAN2017}, Disco-GAN \cite{DBLP:conf/icml/KimCKLK17} and UNIT \cite{liu2017unsupervised} encourage 
the mapping between two domains to be well covered by imposing transitivity: 
the mapping in one direction followed by the mapping in the opposite direction should arrive where it started. 
For this image generation process the main performance measure is either a human-based
quality control or scores that evaluate
the interpretability of the produced images by pre-existing models \cite{Salimans:improvedGAN:NIPS16,CycleGAN2017}.
\vspace{-3mm}
\paragraph{Domain Adaptation}
A widely used 
strategy consists in minimizing
the difference between the source and target distributions \cite{Tzeng_ICCV2015,Sun:CORAL:AAAI16,carlucci2017auto}.
Alternative approaches minimize the errors in target samples reconstruction \cite{WenLi:ECCV2016}
or impose a consistency condition so that neighboring target samples assigned to different labels
are penalized proportionally to their similarity \cite{TRUDA-NIPS16_savarese}. Very recently, \cite{haeusser17}
proposed to enforce associations between source and target samples of the same ground truth or predicted class, while \cite{saito2017asymmetric} assigned pseudo-labels to target samples using an asymmetric tri-training method.

Domain invariance can be also treated as a binary classification problem 
through an adversarial loss inspired by GANs, which encourages mistakes in domain prediction \cite{Ganin:DANN:JMLR16}. For all the methods adopting this strategy, the described losses are minimized jointly with the main 
classification objective 
function on the source task, 
guiding the feature learning process towards a 
domain invariant representation. 
Only in \cite{Hoffman:Adda:CVPR17} the two objectives are kept separated and recombined in a second step. 
In \cite{Bousmalis:DSN:NIPS16} the feature components that differentiate two domains are modeled separately 
from those shared among them.

\vspace{-3mm}
\paragraph{Image Generation for Domain Adaptation}
In the first style transfer methods \cite{Gatys2016a,Johnson2016Perceptual} new images were synthesized 
to maintain a specific content while replicating the style of one or a set of reference images.
Similar transfer approaches have been used to generate images with different visual domains. 
In \cite{Shrivastava:cvpr:17} realistic samples were generated from synthetic images and 
  the produced data could work as training set for a classification model
with good results on real images.
\cite{Bousmalis:Google:CVPR17} proposed a GAN-based approach that adapts source images 
to appear as if drawn from the target domain; the classifier trained 
on such data outperformed several domain adaptation methods by large margins. 
\cite{Taigman2016UnsupervisedCI} introduced a method to generate source images that resemble the target ones, 
with the extra consistency constraint that the same transformation should keep the target samples identical.
All these methods focus on the source-to-target image generation, not considering adding an inverse procedure, from target to source, which we show instead to be beneficial.

\section{Method}
\label{sec:method}
\paragraph{Model}
We focus on unsupervised cross domain classification. Let us start from a dataset $\bm{X}_s=\{\bm{x}^i_s,y^i_s\}_{i=0}^{N_s}$
drawn from a labeled source domain $\mathcal{S}$, and a dataset $\bm{X}_t=\{\bm{x}^j_t\}_{j=0}^{N_t}$ from a different 
unlabeled target domain $\mathcal{T}$, sharing the same set of categories.  
The task is to maximize the classification accuracy on $\bm{X}_t$ while training on $\bm{X}_s$.
To reduce the domain gap, we propose to adapt the source images such that they appear as sampled from 
the target domain by training a generator model $G_{st}$ that maps any source samples 
$\bm{x}^i_s$ to its target-like version $\bm{x}^i_{st}=G_{st}(\bm{x}^i_s)$ defining the set 
$\bm{X}_{st}=\{\bm{x}^i_{st},y^i_s\}_{i=0}^{N_s}$ (see Figure \ref{fig:sbadagan}, bottom row).
The model is also augmented with a discriminator $D_{t}$ and a classifier $C_{t}$. 
The former takes as input the target images $\bm{X}_t$ and target-like source transformed images $\bm{X}_{st}$, 
learning to recognize them as two different sets. The latter takes as input each of 
the transformed images $\bm{x}^i_{st}$ and learns to assign its task-specific label $y^i_s$. 
During the training procedure for this model, information about the domain recognition likelihood 
produced by $D_{t}$ is used adversarially to guide and optimize the performance of the generator $G_{st}$. 
Similarly, the generator also benefits from backpropagation in the classifier training procedure.

Besides the source-to-target transformation, we also consider the inverse target-to-source
direction by using a symmetric architecture (see Figure \ref{fig:sbadagan}, top row).
Here any target image $\bm{x}^j_t$ is given as input to a generator model $G_{ts}$ transforming it to its 
source-like version $\bm{x}^j_{ts}=G_{ts}(\bm{x}^j_t)$, defining the set $\bm{X}_{ts}=\{\bm{x}^j_{ts}\}_{j=0}^{N_t}$. 
As before, the model is augmented with a discriminator $D_{s}$ which takes as input both $\bm{X}_{ts}$ and 
$\bm{X}_s$ and learns to recognize them as two different sets, adversarially helping the generator. 

Since the target images are unlabeled, no classifier can be trained in the target-to-source
direction as a further support for the generator model. We overcome this issue by \emph{self-labeling}
(see Figure \ref{fig:sbadagan}, blue arrow).
The original source data $\bm{X}_s$ is used to train a classifier $C_{s}$.
Once it has reached convergence, we apply the learned model to annotate each of the source-like
transformed target images $\bm{x}^j_{ts}$. These samples, with the assigned pseudo-labels 
$y^j_{t_{self}}=\argmax_y(C_{s}(G_{ts}(\bm{x}^j_t))$, 
are then used transductively as input to $C_{s}$ while information about the performance of the model 
on them is backpropagated to guide and improve the generator $G_{ts}$.
Self-labeling has a long track record of success for domain adaptation: it proved to be effective 
both with shallow models \cite{BruzzonePAMI, HabrardPS13, Morvant15}, as well as with the most recent 
deep architectures \cite{TRUDA-NIPS16_savarese,Tzeng_ICCV2015,saito2017asymmetric}.
In our case the classification loss on pseudo-labeled samples is combined with our other losses, which 
helps making sure we move towards the optimal solution: in case of a moderate domain shift, the 
correct pseudo-labels help to regularize the learning process, while in case of large domain shift, 
the possible mislabeled samples do not hinder the performance (see Sec. \ref{subsec:ablation}
for a detailed discussion on the experimental results).

Finally, the symmetry in the source-to-target and target-to-source transformations is enhanced by 
aligning the two generator models such that, when used in sequence, they bring a sample back to its
starting point. Since our main focus is classification, we are interested in preserving the 
class identity of each sample rather than its overall appearance. Thus, instead of a standard image-based
reconstruction condition we introduce a \emph{class consistency} condition (see Figure \ref{fig:sbadagan}, red arrows).
Specifically, we impose that any source image $\bm{x}^i_s$ adapted to the target domain through $G_{st}(\bm{x}^i_s)$ 
and transformed back towards the source domain through $G_{ts}(G_{st}(\bm{x}^i_s))$ is correctly classified by 
$C_{s}$. This condition helps by imposing a further joint optimization of the two generators. 
\vspace{-3mm}
\paragraph{Learning}
Here we formalize the description above. To begin with, we specify that the generators
take as input a noise vector $\bm{z}\in \mathcal{N}(0,1)$ besides the images, this allows
some extra degree of freedom to model external variations. We also better define
the discriminators as $D_{s}(\bm{x})$, $D_{t}(\bm{x})$ and the classifiers 
as $C_{s}(\bm{x})$, $C_{t}(\bm{x})$. Of course each of these models depends from its parameters 
but we do not explicitly indicate them to simplify the notation. For the same reason
we also drop the superscripts $i,j$. 

The source-to-target part of the network optimizes the following objective function:
\begin{equation}
\min_{{G_{st}}, {C_{t}}} \max_{{D_{t}}} ~~ \alpha \mathcal{L}_{D_t}(D_{t},G_{st}) + 
										   \beta \mathcal{L}_{C_t}(G_{st},C_{t})~,
\end{equation}
where the classification loss $\mathcal{L}_{C_t}$ is a standard \emph{softmax cross-entropy} 
\begin{align}
\mathcal{L}_{C_t}(G_{st},C_{t}) = \mathbb{E}_{\substack{\{\bm{x}_s, \bm{y}_s\}\sim \mathcal{S}\\\bm{z}_s\sim noise}}[{-\bm{y}_{s}}\cdot\log({\hat{\bm{y}}_s})]~,
\label{eq:class}
\end{align}
evaluated on the source samples transformed by the generator $G_{st}$, so that 
$\hat{\bm{y}}_s=C_{t}(G_{st}(\bm{x}_s, \bm{z}_s))$ and $\bm{y}_s$  is the one-hot encoding of the class label $y_s$. 
For the discriminator, instead of the less robust binary cross-entropy,
we followed \cite{mao2016multi} and chose a \emph{least square} loss:
\begin{align}
\mathcal{L}_{D_{t}}(D_{t},G_{st})  = & \mathbb{E}_{\bm{x}_t \sim T}[(D_{t}(\bm{x}_t) - 1)^2 ] + \nonumber \\
                                     & \mathbb{E}_{\substack{\bm{x}_s \sim S\\\bm{z}_s\sim noise}}[(D_{t}(G_{st}(\bm{x}_s, \bm{z}_s)))^2]~. 
                                     \label{eq:discr}
\end{align}

The objective function for the target-to-source part of the network is:
\begin{align}
\min_{{G_{ts}}, {C_{s}} } \max_{{D_{s}}} ~~ & \gamma \mathcal{L}_{D_s}(D_{s},G_{ts}) + \nonumber \\
                                            & \mu    \mathcal{L}_{C_s}(C_{s}) + \eta   \mathcal{L}_{self}(G_{ts},C_{s})~,
\end{align}
where the discriminative loss is analogous to eq. (\ref{eq:discr}), while the
classification loss is analogous to eq. (\ref{eq:class}) but evaluated on the original source samples
with $\hat{\bm{y}}_s = C_{s}(\bm{x}_s)$, 
thus it neither has any dependence on the generator that transforms the target samples $G_{ts}$, 
nor it provides feedback to it.
The \emph{self} loss is again a classification softmax cross-entropy: 
\begin{align}
\mathcal{L}_{self}(G_{ts},C_{s}) = \mathbb{E}_{\substack{\{\bm{x}_t, \bm{y}_{t_{self}}\} \sim \mathcal{T}\\\bm{z}_t\sim noise}}[-\bm{y}_{t_{self}}\cdot \log({\hat{\bm{y}}_{t_{self}}})]~.
\end{align}
where $\hat{\bm{y}}_{t_{self}} = C_{s}(G_{ts}(\bm{x}_t,\bm{z}_t))$ and $\bm{y_{t_{self}}}$ is the one-hot vector
encoding of the assigned label ${y_{t_{self}}}$. 
This loss back-propagates to the generator $G_{ts}$ which is encouraged to preserve the annotated category 
within the transformation.

Finally, we developed a novel \textit{class consistency} loss by minimizing the error of the classifier
$C_{s}$ when applied on the concatenated transformation of $G_{ts}$ and $G_{st}$ to produce
$\hat{\bm{y}}_{cons}= (C_{s}(G_{ts}(G_{st} (\bm{x}_s, \bm{z}_s), \bm{z}_t)))$:
\begin{align}
\mathcal{L}_{cons}(G_{ts},G_{st},C_{s})= 
\mathbb{E}_{\substack{\{\bm{x}_s, \bm{y}_s\}\sim S\\\bm{z}_s,\bm{z}_t \sim noise}}[{-\bm{y}_s}\cdot\log({\hat{\bm{y}}_{cons}})]~.
\end{align}
This loss has the important role of aligning the generators in the two directions and strongly 
connecting the two main parts of our architecture. 

By collecting all the presented parts, we conclude with the complete SBADA-GAN loss: 
\begin{align}
\mathcal{L}_{SBADA-GAN}(G_{st},G_{ts},C_{s},C_{t},D_{s},D_{t})= ~~~~~~~~~~~~~~\nonumber\\
\alpha \mathcal{L}_{D_{t}}+ \beta \mathcal{L}_{C_{t}}+ \gamma \mathcal{L}_{D_{s}} + \mu \mathcal{L}_{C_{s}}+
\eta \mathcal{L}_{self} + \nu \mathcal{L}_{cons}~.                              
\end{align}                            
Here $(\alpha, \beta, \gamma,\mu,\eta,\nu)\geq0$ are weights that control the interaction of the loss terms.  
While the combination of six different losses might appear daunting, it is not unusual \cite{Bousmalis:DSN:NIPS16}.
Here, it stems from the symmetric bi-directional nature of the overall architecture. Indeed each 
directional branch has three losses as it is custom practice in the GAN-based domain adaptation literature \cite{Hoffman:Adda:CVPR17,Bousmalis:Google:CVPR17}. 
Moreover, the ablation study reported in Sec. \ref{subsec:ablation} indicates that the system is remarkably robust to 
changes in the hyperparameter values.
\begin{figure}[tb]
\includegraphics[width=\textwidth]{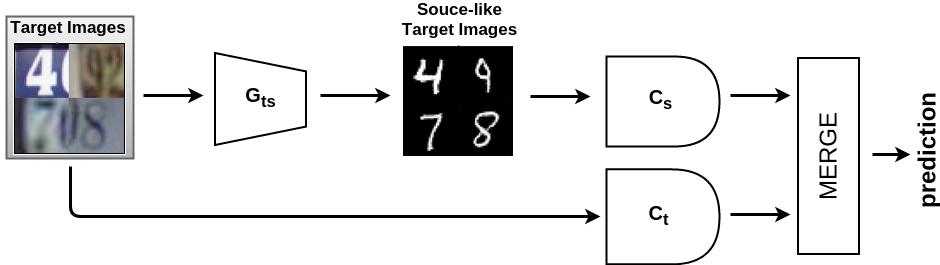}
{\caption{SBADA-GAN, test: the two pre-trained classifiers
are applied respectively on the target images and on the transformed source-like target images.
Their outputs are linearly combined for the final prediction.}\label{fig:cycada_like_test}}
\vspace{-3mm}
\end{figure}
\vspace{-2mm}
\paragraph{Testing}
The classifier $C_{t}$ is trained on $\bm{X}_{st}$ generated 
images that mimic the target domain style, and is then tested on the original target samples $\bm{X}_t$.
The classifier $C_{s}$ is trained on $\bm{X}_s$ source data, and then tested on 
$\bm{X}_{ts}$ samples, that are the target images modified to mimic the source domain style.
These classifiers make mistakes of different type assigning also a different confidence rank to 
each of the possible labels. 
Overall the two classification models complement each other. 
We take advantage of this with a simple ensemble method $~\sigma C_s(G_{ts}(\bm{x}_t,\bm{z}_t)) + \tau C_t(\bm{x}_t)~$ which linearly 
combines their probability output, providing a further gain in performance.
A schematic illustration of the testing procedure is shown in Figure \ref{fig:cycada_like_test}.
We set the combination weights $\sigma,\tau$ through cross validation (see Sec. \ref{subsec:implem} for further details).

\section{Evaluation}
\label{sec:evaluation}
\subsection{Datasets and Adaptation Scenarios}
We evaluate SBADA-GAN on several unsupervised adaptation 
scenarios\footnote{The chosen experimental settings match the ones used in most of 
the previous work involving GAN \cite{Bousmalis:Google:CVPR17,WenLi:ECCV2016,sankaranarayanan2017generate} 
and domain adaptation \cite{Sun:CORAL:AAAI16,Tzeng_ICCV2015,Ganin:DANN:JMLR16,Bousmalis:DSN:NIPS16,Liu:coGAN:NIPS16,Hoffman:Adda:CVPR17}
so we can provide a large benchmark against many methods. 
The standard Office dataset is not considered here, due to the issues clearly explained in \cite{Bousmalis:DSN:NIPS16}, 
specifically in section B of the supplementary material. All the works that show experiments 
on Office exploit networks pre-trained on Imagenet which means involving an extra source 
domain in the task. In our case the network is always trained from scratch on the 
available data of the specific experiment.}, considering the following widely used digits datasets and settings:
\begin{description}[wide=0\parindent]
\item[MNIST $\rightarrow$ MNIST-M:] MNIST~\cite{lecun1998gradient} contains centered, $28\times28$ pixel, grayscale 
images of single digit numbers on a black background, while MNIST-M~\cite{Ganin:DANN:JMLR16} 
is a variant where the  background is substituted by a randomly extracted patch obtained from color photos of BSDS500~\cite{arbelaez2011contour}.
We follow the evaluation protocol of \cite{Bousmalis:DSN:NIPS16,Bousmalis:Google:CVPR17,Ganin:DANN:JMLR16}.
\item[MNIST $\leftrightarrow$ USPS:]USPS~\cite{friedman2001elements} is a digit dataset automatically scanned from 
envelopes by the U.S. Postal Service containing a total of 9,298 $16\times16$ pixel grayscale samples. 
The images are centered, normalized and show a broad range of font styles. 
We follow the evaluation protocol of \cite{Bousmalis:Google:CVPR17}.
\item[SVHN $\leftrightarrow$ MNIST:] SVHN~\cite{netzer2011reading} is the challenging real-world Street View House 
Number dataset, much larger in scale than the other considered datasets. It contains over $600k$ $32\times32$ pixel 
color samples. Besides presenting a great variety of styles (in shape and texture), images from this dataset 
often contain extraneous numbers in addition to the labeled, centered one. Most previous works simplified the 
data by considering a grayscale version, instead we apply our method to the original RGB images. Specifically 
for this experiment we resize the MNIST images to $32\times32$ pixels and use the protocol by 
\cite{Bousmalis:DSN:NIPS16,WenLi:ECCV2016}.
\end{description}

We also test SBADA-GAN on a traffic sign scenario.\\
\textbf{Synth Signs $\rightarrow$ GTSRB:} the Synthetic Signs collection~\cite{Moiseev:2013} contains $100k$ 
samples of common street signs obtained from Wikipedia and artificially transformed to simulate various 
imaging conditions. The German Traffic Signs Recognition Benchmark (GTSRB, \cite{be549cb97c19415788025989a4886d86})
consists of $51,839$ cropped images of German traffic signs.
Both databases contain samples from $43$ 
classes, thus defining a larger classification task than
that on the $10$ digits.  For the experiment we adopt the protocol proposed in \cite{haeusser17}.

\begin{table*}[tbp]
\begin{tabu}{@{}l@{}cccccc}
\hline
\rowfont{\scriptsize}
& \textbf{MNIST$\rightarrow$ USPS}  & \textbf{USPS$\rightarrow$MNIST} 
& \textbf{MNIST$\rightarrow$MNIST-M} & \textbf{SVHN$\rightarrow$MNIST} 
& \textbf{MNIST$\rightarrow$SVHN} & \textbf{Synth Signs$\rightarrow$GTSRB} \\
\hline
\rowfont{\small}Source Only                 			& 78.9   			& 57.1 $\pm$ 1.7    & 63.6      & 60.1 $\pm$ 1.1	& 26.0 $\pm$ 1.2 	& 79.0 \\
\hline
\rowfont{\small}CORAL \cite{Sun:CORAL:AAAI16} 				& 81.7      		&  -                 & 57.7      & 63.1        	& -			& 86.9 \\
\rowfont{\small}MMD \cite{Tzeng_ICCV2015}     				& 81.1      		&  -                 & 76.9      & 71.1         & -         & 91.1 \\
\rowfont{\small}DANN \cite{Ganin:DANN:JMLR16} 				& 85.1      		& 73.0 $\pm$ 2.0    & 77.4      & 73.9          & 35.7      & 88.7 \\
\rowfont{\small}DSN \cite{Bousmalis:DSN:NIPS16}				& 91.3      		&  -                & 83.2      & 82.7          & -         & 93.1 \\
\rowfont{\small}CoGAN \cite{Liu:coGAN:NIPS16}  				& 91.2      		& 89.1 $\pm$ 0.8    & 62.0      & not conv.     &  -     & -  \\
\rowfont{\small}ADDA \cite{Hoffman:Adda:CVPR17} 			& 89.4 $\pm$ 0.2    & 90.1 $\pm$ 0.8    & -     & 76.0 $\pm$ 1.8   & -     & - \\
\rowfont{\small}DRCN \cite{WenLi:ECCV2016}      			& 91.8 $\pm$ 0.1   & 73.7 $\pm$ 0.1   & -      & 82.0 $\pm$ 0.2  & 40.1 $\pm$ 0.1  & - \\
\rowfont{\small}PixelDA \cite{Bousmalis:Google:CVPR17}      & 95.9             & -             & 98.2      &  -              & -            & - \\
\rowfont{\small}DTN \cite{Taigman2016UnsupervisedCI} & -    & -                & -            &  84.4      & -            & - \\
\rowfont{\small}TRUDA \cite{TRUDA-NIPS16_savarese} & -      & -        & 86.7                 & 78.8       & 40.3    & - \\
\rowfont{\small}ATT \cite{saito2017asymmetric} & -      & -        & 94.2                 & 86.2       & 52.8    & 96.2 \\
\rowfont{\small}UNIT \cite{liu2017unsupervised} & 95.9      & 93.5        & -                 & 90.5       & -    & - \\
\rowfont{\small}DA$_{ass}$ fix. par. \cite{haeusser17}     & -         & -             & 89.5           & 95.7             &  -           & 82.8  \\
\rowfont{\small}DA$_{ass}$ \cite{haeusser17}               & -         & -             & 89.5           & \textbf{97.6}             &  -          & \textbf{97.7} \\
\hline
\rowfont{\small}\text{Target Only}            			& 96.5          & 99.2 $\pm$ 0.1    & 96.4     & 99.5     &  96.7     & 98.2        \\
\hline
\rowfont{\small}SBADA-GAN $C_t$          & 96.7         & 94.4         & 99.1           & 72.2             & 59.2 &   95.9      \\
\rowfont{\small}SBADA-GAN $C_s$         & 97.1         & 87.5         & 98.4           & 74.2              & 50.9  &  95.7     \\
\rowfont{\small}SBADA-GAN              & \textbf{97.6} & \textbf{95.0}         & \textbf{99.4}    & 76.1    & \textbf{61.1} & 96.7 \\
\hline
\hline
\rowfont{\small}GenToAdapt \cite{sankaranarayanan2017generate} & 92.5 $\pm$ 0.7    & 90.8 $\pm$ 1.3  & -       & 84.7 $\pm$ 0.9   & 36.4 $\pm$ 1.2   &- \\
\rowfont{\small}CyCADA \cite{cycada} & 94.8 $\pm$ 0.2            & 95.7  $\pm$ 0.2  & -       & 88.3 $\pm$ 0.2  &  -   &- \\
\rowfont{\small}Self-Ensembling \cite{visdawinners} & 98.3 $\pm$ 0.1  & 99.5 $\pm$ 0.4   & -       & 99.2 $\pm$ 0.3  &  42.0 $\pm$ 5.7   & 98.3 $\pm$ 0.3 \\
\hline
\end{tabu}
\caption{
Comparison against previous work.
SBADA-GAN $C_t$ reports the accuracies produced by the classifier trained in the target domain space. 
Similarly, SBADA-GAN $C_s$ reports the results produced by the classifier trained in the source domain space and tested
on the target images mapped to this space. 
SBADA-GAN reports the results obtained by a weighted 
combination of the softmax outputs of these two classifiers. 
Note that all competitors convert SVHN to grayscale, while we deal with the more complex original RGB version. 
The last three rows report results from online available pre-print papers.}
\label{table:results}
\end{table*}

\subsection{Implementation details}
\label{subsec:implem}
We composed SBADA-GAN starting from two symmetric GANs, each with an architecture\footnote{See all the model 
details in the appendix.} analogous to that used for the PixelDA model \cite{Bousmalis:Google:CVPR17}.


The model is coded in python and we ran all our experiments in the Keras framework \cite{chollet2017} 
(code will be made available upon acceptance). We use the ADAM \cite{kingma2014adam}
optimizer with learning rates for the discriminator and the generator both set to $10^{-4}$.
The batch size is set to $32$ and we train the model for $500$ epochs not noticing any overfitting, which suggests that further epochs might be beneficial.
The $\alpha$ and $\gamma$ loss weights (discriminator losses) are set to $1$, 
$\beta$ and $\mu$ (classifier losses) are set to $10$, 
to prevent that generator from indirectly 
switching labels (for instance, transform $7$'s into $1$'s). 
The class consistency loss weight $\nu$ is set to $1$.
All training procedures start with the self-labeling loss weight, $\eta$, set to zero, as 
this loss hinders convergence until the classifier is fully trained. After the model converges 
(losses stop oscillating, usually after $250$ epochs) $\eta$ is set to $1$ to further 
increase performance. 
Finally the parameters to combine the classifiers at test time are $\sigma \in [0,0.1,0.2, \ldots, 1]$ 
and $\tau=(1-\sigma)$ chosen on a validation set of 1000 random samples from the target in
each different setting.

\subsection{Quantitative Results}
Table \ref{table:results} shows results on our six evaluation settings.
The top of the table reports
results
by thirteen competing 
baselines published over the last two years.
The Source-Only and Target-Only
rows contain reference results corresponding to the na\"ive no-adaptation case and to the
target fully supervised case. 
For SBADA-GAN, besides the full method, we also report the accuracy obtained by the separate 
classifiers (indicated by $C_s$ and $C_t$) before the linear combination.
The last three rows 
show results that 
appeared recently in pre-prints available online.

SBADA-GAN improves over the state of the art in four out of six settings. 
In these cases the advantage with respect to its competitors is already visible in the 
separate $C_s$ and $C_t$ results and it increases when considering the full combination procedure. 
Moreover, the gain in performance of SBADA-GAN reaches up to $+8$ percentage points 
in the MNIST$\rightarrow$SVHN experiment. This setting was disregarded in many previous 
works: differently from its
inverse SVHN$\rightarrow$MNIST, it requires a difficult adaptation from the 
grayscale handwritten digits domain to the widely variable and colorful street view house number domain.
Thanks to its bi-directionality, SBADA-GAN leverages on the inverse target to source mapping 
to produce highly accuracy results.

Conversely, 
in the SVHN$\rightarrow$MNIST case SBADA-GAN ranks eighth 
out of the thirteen baselines in terms of performance. 
Our accuracy is on par with ADDA's \cite{Hoffman:Adda:CVPR17}: the two approaches
share the same classifier architecture, although the number of fully-connected neurons of 
SBADA-GAN is five time lower. Moreover, compared to DRCN~\cite{WenLi:ECCV2016}, the classifiers
of SBADA-GAN are shallower with a reduced number of convolutional layers. 
Overall here SBADA-GAN 
suffers of some
typical drawbacks of GAN-based domain adaptation methods: although the style of a domain
can be easily transferred in the raw pixel space, the generative process does not have 
any explicit constraint on reducing the overall data distribution shift as instead done 
by the alternative feature-based domain adaptation approaches. Thus, methods like 
DA$_{ass}$~\cite{haeusser17}, DTN~\cite{Taigman2016UnsupervisedCI} and DSN~\cite{Bousmalis:DSN:NIPS16} 
deal better 
with the large domain gap of the SVHN$\rightarrow$MNIST setting.

Finally, in the Synth Signs $\rightarrow$ GTSRB experiment, 
SBADA-GAN is just slightly worse than DA$_{ass}$, but 
outperforms all the other competing methods.
The comparison remains in favor of SBADA-GAN when considering that 
its performance is robust to hyperparameter variations (see Sec. \ref{subsec:ablation} for more details), 
while the performance of DA$_{ass}$ drops significantly in case of not tuned, 
pre-defined fixed parameters.

\subsection{Qualitative Results}
\begin{figure}[tb]
    \centering
   \begin{subfigure}[b]{0.45\textwidth}
        \includegraphics[width=\textwidth]{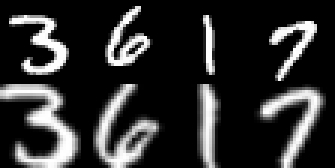}
        \caption{MNIST to USPS}
        \label{fig:mnist_to_usps}
    \end{subfigure}
        \begin{subfigure}[b]{0.45\textwidth}
        \includegraphics[width=\textwidth]{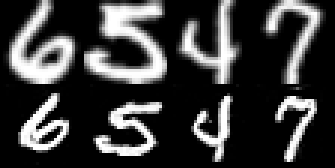}
        \caption{USPS to MNIST}
        \label{fig:usps_to_mnist}
    \end{subfigure}
    \begin{subfigure}[b]{0.45\textwidth}
        \includegraphics[width=\textwidth]{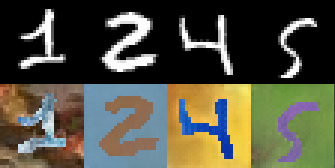}
        \caption{MNIST to MNIST-M}
        \label{fig:mnist_to_mnistM}
    \end{subfigure}
     \begin{subfigure}[b]{0.45\textwidth}
        \includegraphics[width=\textwidth]{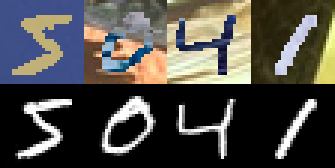}
        \caption{MNIST-M to MNIST}
        \label{fig:mnistmM_to_mnist}
    \end{subfigure}

     \begin{subfigure}[b]{0.45\textwidth}
        \includegraphics[width=\textwidth]{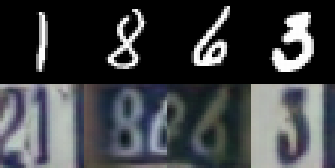}
        \caption{MNIST to SVHN}
        \label{fig:mnist_to_svhn}
    \end{subfigure}    
         \begin{subfigure}[b]{0.45\textwidth}
        \includegraphics[width=\textwidth]{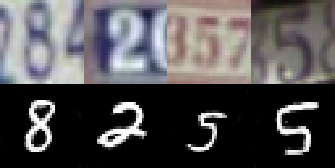}
        \caption{SVHN to MNIST}
        \label{fig:svhn_to_mnist}
    \end{subfigure}

         \begin{subfigure}[b]{0.45\textwidth}
        \includegraphics[width=\textwidth]{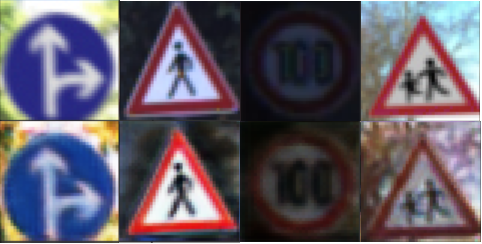}
        \caption{Synth S. to GTSRB}
        \label{fig:synth_to_gtsrb}
    \end{subfigure}  
    \begin{subfigure}[b]{0.45\textwidth}
        \includegraphics[width=\textwidth]{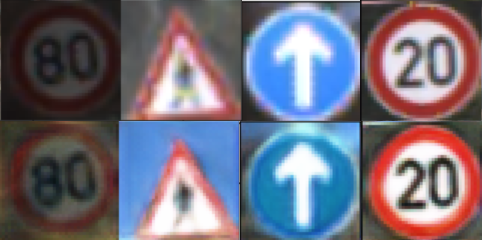}
        \caption{GTSRB to Synth S.}
        \label{fig:gtsrb_to_synth}
    \end{subfigure}  
\caption{Examples of generated digits: we show the image transformation
    from the original domain to the paired one as indicated under every sub-figure.
    For each of the (a)-(h) cases, the original/generated images are in the top/bottom row.
    }\label{fig:examples_best}
    \vspace{-2mm}
\end{figure}

To complement the quantitative evaluation, we look at
the quality of the images generated by SBADA-GAN. First,
we see from Figure \ref{fig:examples_best}
how the generated images
actually mimic the style of the chosen domain, even when going from the simple MNIST
digits to the SVHN colorful house numbers.

Visually inspecting the data distribution before and after domain mapping defines a second qualitative
evaluation metric. We use t-SNE~\cite{maaten2008visualizing} to project the data from their 
raw pixel space to a simplified 2D embedding. Figure \ref{fig:TSNE} shows such 
visualizations and indicates that the transformed dataset tends to replicate faithfully
the distribution of the chosen final domain.

A further measure of the quality of the SBADA-GAN generators 
comes from the
diversity of the produced images. Indeed, a well-known failure mode of GANs is that the 
generator may collapse and output a single prototype that maximally fools the discriminator.
To evaluate the diversity of samples generated by SBADA-GAN we choose 
the Structural Similarity (SSIM, \cite{Wang:2004:IQA}),
a measure that correlates well with the human perceptual similarity judgments. 
Its values range between $0$ and $1$ with higher values 
corresponding to more similar images. We follow the same procedure used in 
\cite{pmlr-v70-odena17a} by randomly choosing $1000$ pairs of generated images within a given class. 
We also repeat the evaluation over all the classes and calculate the average results.
Table \ref{table:data_ssim} shows the results of the mean SSIM metric
and indicates that the SBADA-GAN generated images not only mimic the same style, but also
successfully reproduce the variability of a chosen domain.

\begin{table}[tb]
\centering
\begin{tabu}{lcccc}
\hline
\rowfont{\scriptsize} Setting & S   & T map to S & S map to T & T   \\
\hline
\rowfont{\scriptsize} MNIST $\rightarrow$ USPS    & $0.206$  &   $0.219$      &   $0.106$      & $0.102$  \\
\rowfont{\scriptsize} MNIST $\rightarrow$ MNIST-M & $0.206$  &   $0.207$      &  $0.035$   & $0.032$  \\     
\rowfont{\scriptsize} MNIST $\rightarrow$ SVHN    & $0.206$  &   $0.292$      &  $0.027$       & $0.012$  \\
\rowfont{\scriptsize} Synth S. $\rightarrow$ GTSRB       & $0.105$  &   $0.136$      &  $0.128$       & $0.154$  \\
\hline
\end{tabu}
\caption{Dataset mean SSIM: this measure of data variability suggests 
that our method successfully generates images with not only the same style of a chosen domain, 
but also similar perceptual variability.}
\label{table:data_ssim}
\end{table}
\begin{figure}[tb]
    \centering
    \vspace{-0.5em}
    \begin{subfigure}[b]{0.49\textwidth}
        \includegraphics[width=\textwidth]{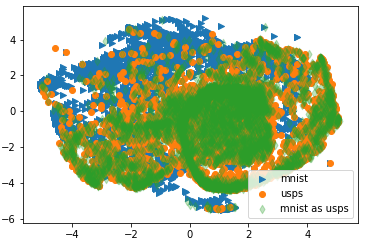}
        \caption{MNIST to USPS}
        \label{fig:mnist_usps_1}
    \end{subfigure}
    \begin{subfigure}[b]{0.49\textwidth}
        \includegraphics[width=\textwidth]{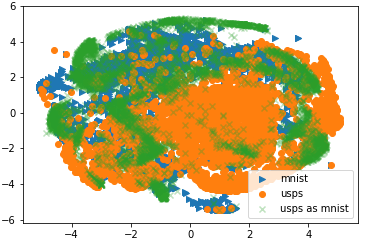}
        \caption{USPS to MNIST}
        \label{fig:mnist_usps_2}
    \end{subfigure}
    \vspace{-0.5em}
     \begin{subfigure}[b]{0.49\textwidth}
        \includegraphics[width=\textwidth]{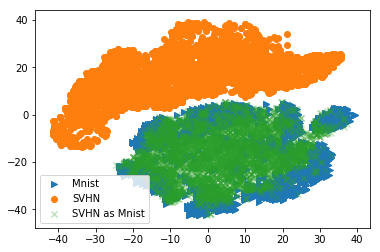}
        \caption{SVHN to MNIST}
        \label{fig:mnist_svhn_1}        
    \end{subfigure}
        \begin{subfigure}[b]{0.49\textwidth}
        \includegraphics[width=\textwidth]{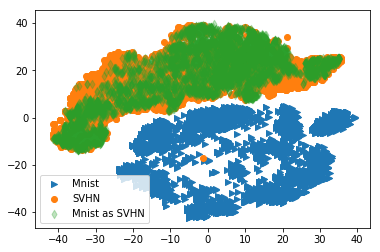}
        \caption{MNIST to SVHN}
        \label{fig:mnist_svhn_2}       
    \end{subfigure}
    \caption{t-SNE visualization of source, target and source mapped to target images. 
    Note how the mapped source covers faithfully the target space both in the (a),(b)
    case with moderated domain shift and in the more challenging (c),(d) setting.}\label{fig:TSNE}
\end{figure}

\subsection{Ablation and Robustness Study}
\label{subsec:ablation}
To clarify the role of each component in SBADA-GAN we go back to its
core source-to-target single GAN module and analyze the effect of adding all the other 
model parts. Specifically we start by adding the symmetric target-to-source GAN model.
These two parts are then combined and the domain transformation loop is closed 
by adding the class consistency condition. Finally the model is completed by introducing the 
target self-labeling procedure. We empirically test each of these model reconstruction steps 
on the MNIST$\rightarrow$USPS setting and report the results in Table \ref{table:ablation}.
We see
the gain achieved by progressively adding the different
model components, with the largest advantage obtained by the introduction of self-labeling. 

An analogous boost due to self-labeling is also visible in all the other experimental
settings with the exception of  MNIST$\leftrightarrow$SVHN, where the accuracy remains
unchanged if $\eta$ is equal or larger than zero. A further analysis reveals that here 
the recognition accuracy of the source classifier applied to the source-like transformed target images
is quite low (about $65\%$, while in all the other settings reaches $80-90\%$), thus the 
pseudo-labels cannot be considered reliable. Still, using them does not hinder the 
overall performance.

\begin{table}[t]
\centering
\begin{tabu}{|c|c|c|c|c|c|@{}c@{}|}
\hline
\rowfont{\small}\multicolumn{2}{|c|}{S$\rightarrow$T} & \multicolumn{2}{c|}{T$\rightarrow$S}& Class & Self & \multirow{2}{*}{Accuracy}\\
\rowfont{\small}\multicolumn{2}{|c|}{GAN}      & \multicolumn{2}{c|}{GAN} & Consist. & Label. &\\
\cline{1-6}
\rowfont{\small}
\parbox[t]{2mm}{\rotatebox[origin=c]{90}{$\mathcal{L}_{D_{t}}$}}& 
\parbox[t]{2mm}{\rotatebox[origin=c]{90}{$\mathcal{L}_{C_{t}}$}}& 
\parbox[t]{2mm}{\rotatebox[origin=c]{90}{$\mathcal{L}_{D_{s}}$}}& 
\parbox[t]{2mm}{\rotatebox[origin=c]{90}{$\mathcal{L}_{C_{s}}$}}& 
\parbox[t]{2mm}{\rotatebox[origin=c]{90}{$\mathcal{L}_{cons}$}}& 
\parbox[t]{2mm}{\rotatebox[origin=c]{90}{$\mathcal{L}_{self}$}}& 
~~MNIST$\rightarrow$USPS~~\\						\hline
$\checkmark$ & $\checkmark$ &&&&&94.23 \\
&&$\checkmark$&$\checkmark$&&&91.55 \\
$\checkmark$&$\checkmark$&$\checkmark$&$\checkmark$&&&94.90 \\
$\checkmark$&$\checkmark$&$\checkmark$&$\checkmark$&$\checkmark$&&95.45 \\
$\checkmark$&$\checkmark$&$\checkmark$&$\checkmark$&$\checkmark$&$\checkmark$&97.60\\
\hline
\end{tabu}
\caption{Analysis of the role of each SBADA-GAN component. We ran experiments by turning on
the different losses of the model as indicated by the checkmarks. 
}
\label{table:ablation}
\end{table}

\begin{figure}[tb]
    \centering
    \begin{subfigure}[b]{0.48\textwidth}
        \includegraphics[width=\textwidth]{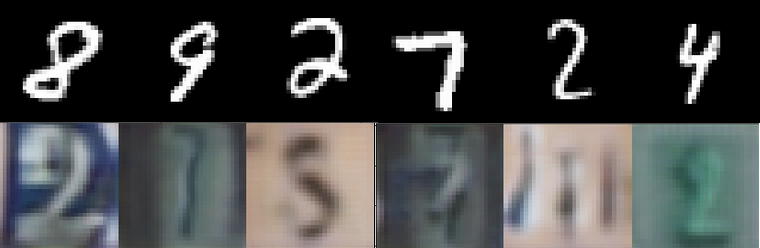}
        \caption{}
        \label{fig:mnistm_to_mnistA_1}
    \end{subfigure}
        \begin{subfigure}[b]{0.48\textwidth}
        \includegraphics[width=\textwidth]{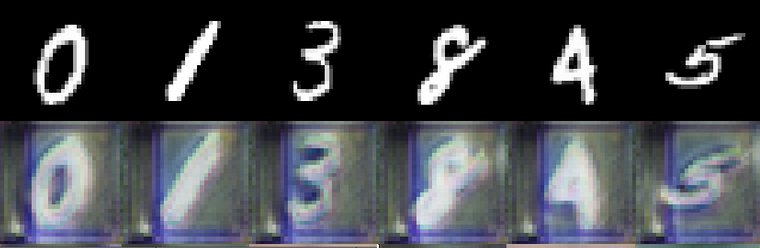}
        \caption{}
        \label{fig:mnistm_to_mnistA_2}
    \end{subfigure}
    \begin{subfigure}[b]{0.48\textwidth}
        \includegraphics[width=\textwidth]{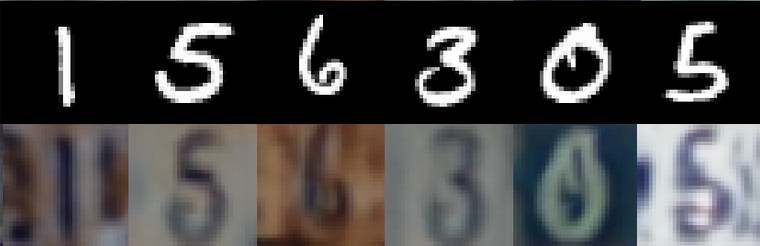}
        \caption{}
        \label{fig:mnistm_to_mnistA_3}
    \end{subfigure}
        \begin{subfigure}[b]{0.48\textwidth}
        \includegraphics[width=\textwidth]{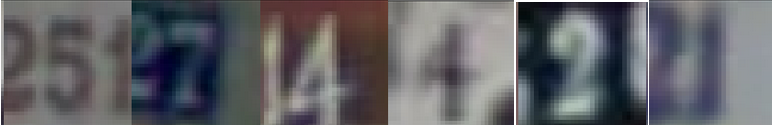}
        \caption{}
        \label{fig:mnistm_to_mnistA_4}
    \end{subfigure}
    \caption{$G_{ts}$ outputs (lower line) and their respective inputs (upper line) obtained with: 
              (a) no consistency loss, (b) image-based cycle consistency loss \cite{CycleGAN2017,DBLP:conf/icml/KimCKLK17}, 
              (c) our class consistency loss. 
              In (d) we show some real SVHN samples as a reference.}\label{fig:rec_comparison}
\vspace{-3mm}
\end{figure}

The crucial effect of the class consistency loss can be better observed by looking 
at the generated images and comparing them with those obtained in two alternative cases: 
setting $\nu=0$, \ie not using any consistency condition between the two generators $G_{st}$ and $G_{ts}$, 
or substituting our class consistency loss with the standard cycle consistency loss \cite{CycleGAN2017,DBLP:conf/icml/KimCKLK17}
based on image reconstruction.
For this evaluation we choose the MNIST$\rightarrow$SVHN case which has the strongest
domain shift and we show the generated images in Figure \ref{fig:rec_comparison}.
When the consistency loss is not activated, the $G_{ts}$ output images are realistic, 
but fail at reproducing the correct input digit and provide misleading information to the classifier. 
On the other hand, using the cycle-consistency loss preserves the input digit but fails in rendering a 
realistic sample in the correct domain style. 
Finally, our class consistency loss allows to preserve the distinct features belonging to a 
category while still leaving enough freedom to the generation process, thus it 
succeeds in both preserving the digits and rendering realistic samples. 

About the class consistency loss, we also note that SBADA-GAN is robust to the specific
choice of the weight $\nu$, given that it is different from zero. Changing it in $[0.1,1,10]$ induces
a maximum variation of $0.6$ percentage points in accuracy over the different settings. 
An analogous evaluation performed on the classification loss weights $\beta$ and $\mu$ reveals that changing them 
in the same range used for $\nu$ causes a maximum overall performance variation of $0.2$ percentage points. 
Furthermore SBADA-GAN is minimally sensitive to the  batch size used: halving it from $32$ to $16$ samples 
while keeping the same number of learning epochs reduces the performance only of about $0.2$ percentage points.
Such robustness is
particularly relevant when 
compared to
competing 
methods. For instance the most 
recent $DA_{ass}$~\cite{haeusser17}  
needs a perfectly balanced source and target distribution of classes in each batch, 
a condition difficult to satisfy in real world scenarios, and halving the originally large batch size 
reduces by $3.5$ percentage points the final accuracy. Moreover, changing the weights of the 
losses that enforce associations 
across domains with a range analogous to that used for the SBADA-GAN parameters induces a 
drop in performance up to $16$ accuracy percentage points\footnote{
More details about these experiments on robustness
are provided in the appendix.}.

\section{Conclusion}
\label{sec:conclusion}
This paper presented SBADA-GAN, an adaptive adversarial domain adaptation architecture that maps 
simultaneously source samples into the target domain and vice versa with the aim to 
learn and use both classifiers at test time. To achieve this, we proposed 
to use self-labeling to regularize the classifier trained on the source, and  we impose a class 
consistency loss that improves greatly the stability of the architecture, as well as the quality 
of the reconstructed images in both domains. 

We explain the success of SBADA-GAN in several ways.
To begin with, thanks to the the bi-directional mapping we avoid deciding a priori which is the best 
strategy for a specific task. Also, the combination of the two network directions improves 
performance providing empirical evidence that they are learning different, complementary features.
Our class consistency loss aligns the image generators, allowing both domain transfers to 
influence each other.  Finally the self-labeling procedure boost the performance in case of moderate 
domain shift without hindering it in case of large domain gaps.

\section*{Acknowledgments}
This work was supported by the ERC grant 637076 - RoboExNovo.

\newpage
\appendix
\section*{Appendix}
\label{sec:appendix}
\section{SBADA-GAN network architecture}
We composed SBADA-GAN starting from two symmetric GANs, each with an architecture
analogous to that used for the PixelDA model.
Specifically 
\begin{itemize}[leftmargin=*]
\item the generators take the form of a convolutional residual network 
with four residual blocks each composed by two convolutional layers with 64 features;
\item the input noise $\bm{z}$ is a vector of $N^z$ elements each sampled from a 
normal distribution $z_i \sim\mathcal{N}(0,1)$. It is fed to a fully connected
layer which transforms it to a channel of the same resolution as that of the 
image, and is subsequently concatenated to the input as an extra channel. 
In all our experiments we used $N^z = 5$;
\item the discriminators are made of two convolutional layers, followed by an average pooling and a 
convolution that brings the discriminator output to a single scalar value;
\item in both generator and discriminator networks, each convolution (with the exception of the last one of 
the generator) is followed by a batch norm layer \cite{ioffe2015batch};
\item the classifiers have exactly the same structure of that in \cite{Bousmalis:Google:CVPR17,Ganin:DANN:JMLR16};
\item as activation functions we used ReLU in the generator and classifier, while we used leaky ReLU (with a $0.2$ slope) 
in the discriminator. 
\item all the input images to the generators are zero-centered and rescaled to $[-0.5,0.5]$. The images produced
by the generators as well as the other input images to the classifiers and and the discriminators are zero-centered
and rescaled to $[-127.5,127.5]$.
\end{itemize}
Thanks to the stability of the SBADA-GAN training protocol, we did not use any injected noise 
into the discriminators and we did not use any dropout layer.

\section{Experimental Settings}

\begin{description}[wide=0\parindent]
\item[MNIST $\rightarrow$ MNIST-M:] 
MNIST has $60k$ images for training. 
As \cite{Bousmalis:Google:CVPR17} we divided it into $50k$ samples for actual training 
and $10k$ for validation. 
All the $60k$ images from the MNIST-M training set were considered
as test set. A subset of $1k$ images and their labels were also used to validate the 
classifier combination weights at test time.
\item[USPS $\rightarrow$ MNIST:] 
USPS has $6,562$ training, $729$ validation, and $2,007$ test images. All of them were resized to 
$28\times28$ pixels. 
The $60k$ training images of MNIST were considered as test set,
with $1k$ samples and their labels also used for validation purposes.
\item[MNIST $\rightarrow$ USPS:] 
even in this case MNIST training images were divided into $50k$ samples for actual 
training and $10k$ for validation. 
We tested on the whole set of $9,298$ images of USPS. 
Out of them, $1k$ USPS images and their labels were also used for validation.
\item[SVHN $\rightarrow$ MNIST:] 
SVHN  contains over $600k$ color images of which $73,257$ samples are used for training
and $26,032$ for validation while the remaining data are somewhat less difficult samples. 
We disregarded this last set and considered only the first two. 
The $60k$ MNIST training samples were considered as test set,
with $1k$ MNIST images and their labels also used for validation.
\item[MNIST $\rightarrow$ SVHN:] 
for MNIST we used again the $50k$/$10k$ training/validation sets. 
The whole set of $99,289$ SVHN samples was considered for testing with 
$1k$ images and their labels also used for validation.
\item[Synth Signs $\rightarrow$ GTSRB:] 
the Synth Signs dataset contains $100k$ images, out of which $90k$ were used for training and $10k$ for validation. The model was tested on the whole GTSRB dataset containing $51,839$ samples 
resized with bilinear interpolation to match the Synth Signs images' size of $40\times40$ pixels. Similarly to the
previous cases, $1k$ GTSRB images and their labels were considered for validation purposes.
\end{description}

\section{Distribution Visualizations}
To visualize the original data distributions and their respective transformations we used t-SNE \cite{maaten2008visualizing}. 
The images were pre-processed by scaling in $[-1,1]$ and we applied PCA for dimensionality reduction 
from vectors with Width$\times$Height elements to $64$ elements. Finally t-SNE with default parameters 
was applied to project data to a 2-dimensional space.

The behavior shown by the t-SNE data visualization presented in the main paper 
extends also for the other experimental settings. We integrate here the
visualization for the MNIST$\rightarrow$MNIST-M case in Figure \ref{fig:TSNE}.
The plots show again a  successful mapping with the generated 
data that cover faithfully the target space.

\begin{figure}[tb]
    \centering
    \vspace{-0.5em}
    \begin{subfigure}[b]{0.49\textwidth}
        \includegraphics[width=\textwidth]{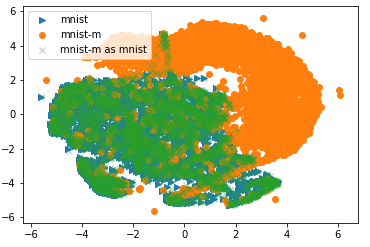}
        \caption{MNIST-M to MNIST}
        \label{fig:mnist_mnistm_1}   
    \end{subfigure}
        \begin{subfigure}[b]{0.49\textwidth}
        \includegraphics[width=\textwidth]{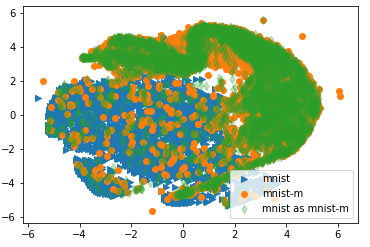}
        \caption{MNIST to MNIST-M}
        \label{fig:mnist_mnistm_2}      
    \end{subfigure}
    \caption{t-SNE visualization of source, target and source mapped to target images. 
    Note how the mapped source covers faithfully the target space in all the settings.}\label{fig:TSNE}
\end{figure}

\section{Robustness experiments}

\begin{figure}[tb]
\vspace{-0.5em}
\begin{tabular}{c@{~~}c}
\includegraphics[width=0.47\textwidth]{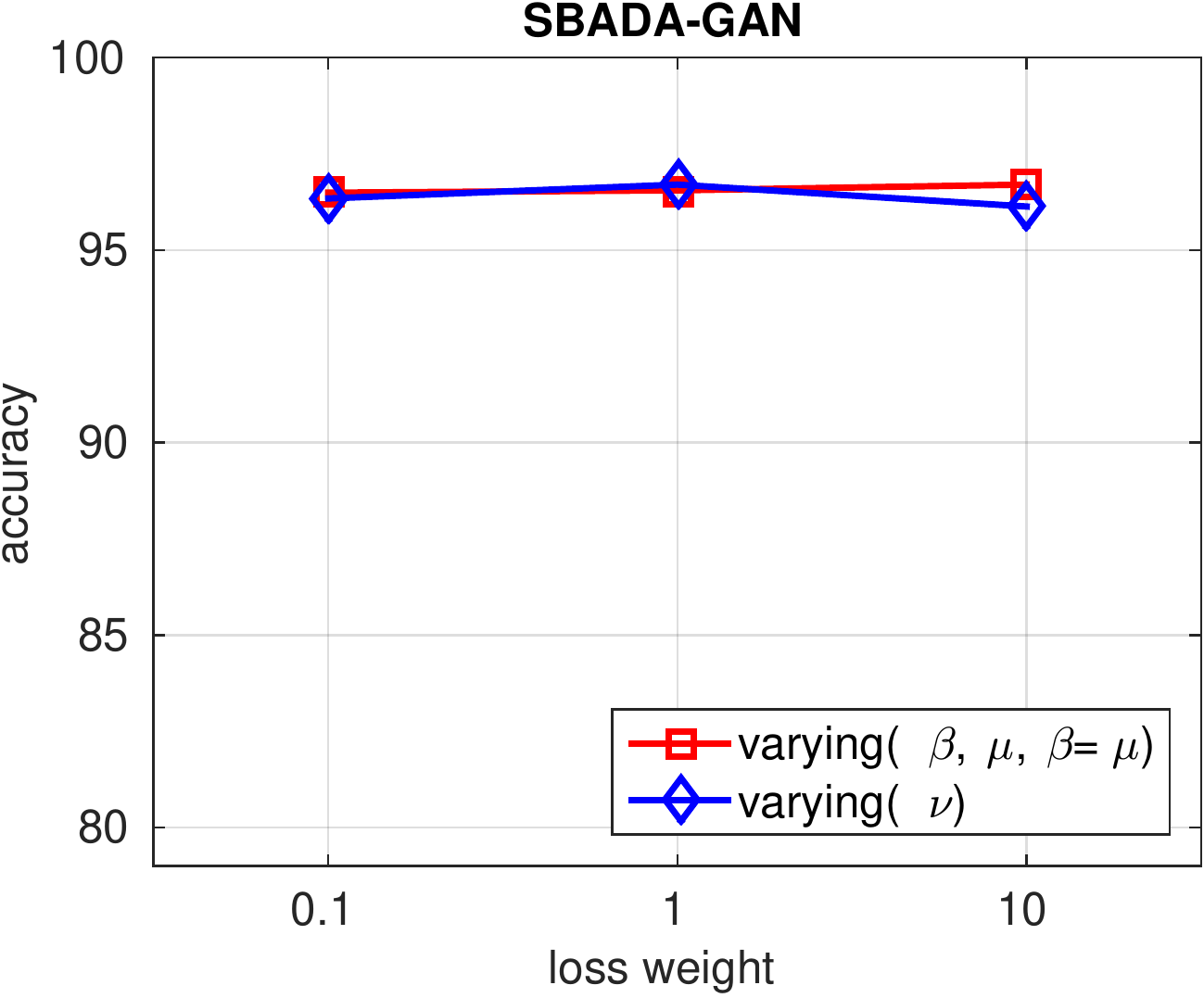}&\includegraphics[width=0.47\textwidth]{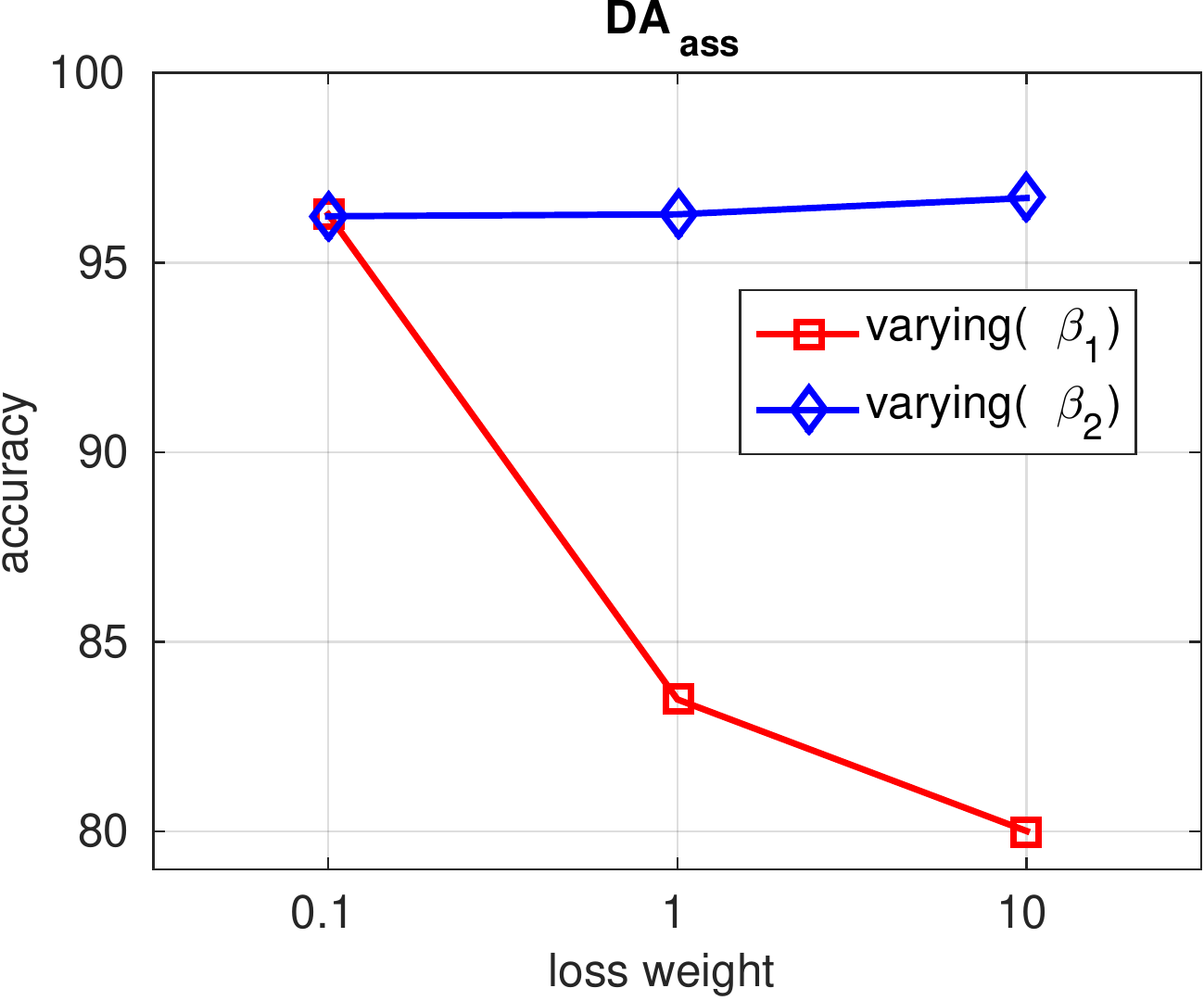}
\end{tabular}
\caption{Behaviour of the SBADA-GAN and DA$_{ass}$ methods when changing their loss weights.
(left) for SBADA-GAN we kept $\alpha=\gamma=1$ and $\eta=1$, 
while we varied alternatively the weights of the classification losses $\beta,\mu$ with $\beta=\mu$ and keeping $\nu=1$, 
or the weight of the class consistency loss $\nu$ while fixing $\beta=\mu=10$.
(right) for the DA$_{ass}$ method we changed the weight of the walker loss $\beta_1$ while keeping that of the
visit loss $\beta_2=0.1$, or alternatively we changed the weight of the visit loss $\beta_2$ while fixing
that of the walker loss $\beta_1=1$. 
}
\label{fig:robustness}
\end{figure}

The experiments about SBADA-GAN robustness to hyperparameters values are described at high level in Section 4.5
of the main paper submission. Here we report on the detailed results obtained on Synth. Signs $\rightarrow$ GTSRB when using SBADA-GAN and the DA$_{ass}$ method \cite{haeusser17}. 

For SBADA-GAN we keep fixed the weights of the discriminative losses $\alpha=\gamma=1$ as well as that of self-labeling
$\eta=1$, while we varied alternatively the weights of the classification losses $\beta,\mu$ or the weight of the
class consistency loss $\nu$ in $[0.1, 1, 10]$. The results plotted in Figure \ref{fig:robustness} (left) show 
that the classification accuracy changes less than 0.2 percentage point. 
Furthermore, we used a batch size of 32 for our experiments and 
when reducing it to 16 the overall accuracy remains almost unchanged (from 96.7 to 96.5). 

DA$_{ass}$ proposes to minimize the difference between the source and target by maximizing
the associative similarity across domains. This is based on the two-step round-trip probability of an imaginary random 
walker starting from a sample $(x^s_i, y_i)$ of the source domain, passing through an unlabeled 
sample of the target domain $(x^t_j)$ and and returning to another source sample $(x^s_k, y_k=y_i)$ 
belonging to the same class of the initial one. This is formalized by first assuming that all the categories
have equal probability both in source and in target, and then measuring the difference between the uniform 
distribution and the two-step probability through the so called \emph{walker loss}. To avoid that only few target 
samples are visited multiple times, a second \emph{visit loss} measures the difference between the uniform 
distribution and the probability of visiting some target samples. 
We tested the robustness of DA$_{ass}$ by using the code provided by its authors and changing the 
loss weights $\beta_1$ for the walker loss and $\beta_2$ for the visit loss in the same range used 
for the SBADA-GAN: [0.1 1 10]. Figure \ref{fig:robustness} (right)
shows that DA$_{ass}$ is particularly sensitive to modifications of the visit loss weights which can
cause a drop in performance of more than 16 percentage points. 
Moreover, the model assumption about the class balance sounds too strict for realistic scenarios: in 
practice DA$_{ass}$ needs every observed data batch to contain an equal number of samples from each category
and reducing the number of samples from 24 to 12 per category causes a drop in performance of more than 4 
percentage points from 96.3 to 92.8. 

To conclude, although GAN methods are generally considered unstable and difficult to train, 
SBADA-GAN results much more robust than a not-GAN approach like DA$_{ass}$ to the loss weights 
hyperparameters and can be trained with small random batches of data while not losing its high 
accuracy performance.

{\small
\bibliographystyle{ieee}
\bibliography{egbib}

\begin{thebibliography}{10}\itemsep=-1pt

\bibitem{cycada}
{CyCADA: Cycle-Consistent Adversarial Domain Adaptation}.
\newblock In {\em Blind Submission, International Conference on Learning
  Representations (ICLR)}, 2018.

\bibitem{visdawinners}
Self-ensembling for visual domain adaptation.
\newblock In {\em Blind Submission, International Conference on Learning
  Representations (ICLR)}, 2018.

\bibitem{arbelaez2011contour}
P.~Arbelaez, M.~Maire, C.~Fowlkes, and J.~Malik.
\newblock Contour detection and hierarchical image segmentation.
\newblock {\em IEEE Trans. Pattern Anal. Mach. Intell.}, 33(5):898--916, 2011.

\bibitem{Bousmalis:Google:CVPR17}
K.~Bousmalis, N.~Silberman, D.~Dohan, D.~Erhan, and D.~Krishnan.
\newblock Unsupervised pixel-level domain adaptation with gans.
\newblock In {\em Computer Vision and Pattern Recognition (CVPR)}, 2017.

\bibitem{Bousmalis:DSN:NIPS16}
K.~Bousmalis, G.~Trigeorgis, N.~Silberman, D.~Krishnan, and D.~Erhan.
\newblock {Domain Separation Networks}.
\newblock In {\em {Neural Information Processing Systems (NIPS)}}, 2016.

\bibitem{BruzzonePAMI}
L.~Bruzzone and M.~Marconcini.
\newblock Domain adaptation problems: A dasvm classification technique and a
  circular validation strategy.
\newblock {\em IEEE Trans. Pattern Anal. Mach. Intell.}, 32(5):770--787, 2010.

\bibitem{carlucci2017auto}
F.~M. Carlucci, L.~Porzi, B.~Caputo, E.~Ricci, and S.~Rota~Bul{\`o}.
\newblock Autodial: Automatic domain alignment layers.
\newblock In {\em International Conference on Computer Vision (ICCV)}, 2017.

\bibitem{chollet2017}
F.~Chollet.
\newblock keras.
\newblock \url{https://github.com/fchollet/keras}, 2017.

\bibitem{friedman2001elements}
J.~Friedman, T.~Hastie, and R.~Tibshirani.
\newblock {\em The elements of statistical learning}, volume~1.
\newblock Springer series in statistics Springer, Berlin, 2001.

\bibitem{Ganin:DANN:JMLR16}
Y.~Ganin, E.~Ustinova, H.~Ajakan, P.~Germain, H.~Larochelle, F.~Laviolette,
  M.~Marchand, and V.~Lempitsky.
\newblock Domain-adversarial training of neural networks.
\newblock {\em J. Mach. Learn. Res.}, 17(1):2096--2030, 2016.

\bibitem{Gatys2016a}
L.~A. Gatys, A.~S. Ecker, and M.~Bethge.
\newblock Image style transfer using convolutional neural networks.
\newblock In {\em Conference on Computer Vision and Pattern Recognition
  (CVPR)}, 2016.

\bibitem{WenLi:ECCV2016}
M.~Ghifary, W.~B. Kleijn, M.~Zhang, D.~Balduzzi, and W.~Li.
\newblock Deep reconstruction-classification networks for unsupervised domain
  adaptation.
\newblock In {\em European Conference on Computer Vision (ECCV)}, 2016.

\bibitem{Goodfellow:GAN:NIPS2014}
I.~Goodfellow, J.~Pouget-Abadie, M.~Mirza, B.~Xu, D.~Warde-Farley, S.~Ozair,
  A.~Courville, and Y.~Bengio.
\newblock Generative adversarial nets.
\newblock In {\em {Neural Information Processing Systems (NIPS)}}. 2014.

\bibitem{HabrardPS13}
A.~Habrard, J.-P. Peyrache, and M.~Sebban.
\newblock Iterative self-labeling domain adaptation for linear structured image
  classification.
\newblock {\em International Journal on Artificial Intelligence Tools}, 22(5),
  2013.

\bibitem{haeusser17}
P.~Haeusser, T.~Frerix, A.~Mordvintsev, and D.~Cremers.
\newblock Associative domain adaptation.
\newblock In {\em International Conference on Computer Vision (ICCV)}, 2017.

\bibitem{ioffe2015batch}
S.~Ioffe and C.~Szegedy.
\newblock Batch normalization: Accelerating deep network training by reducing
  internal covariate shift.
\newblock In {\em International Conference on Machine Learning}, pages
  448--456, 2015.

\bibitem{Johnson2016Perceptual}
J.~Johnson, A.~Alahi, and L.~Fei-Fei.
\newblock Perceptual losses for real-time style transfer and super-resolution.
\newblock In {\em European Conference on Computer Vision (ECCV)}, 2016.

\bibitem{DBLP:conf/icml/KimCKLK17}
T.~Kim, M.~Cha, H.~Kim, J.~K. Lee, and J.~Kim.
\newblock Learning to discover cross-domain relations with generative
  adversarial networks.
\newblock In {\em International Conference on Machine Learning, {ICML}}, pages
  1857--1865, 2017.

\bibitem{kingma2014adam}
D.~Kingma and J.~Ba.
\newblock Adam: A method for stochastic optimization.
\newblock In {\em International Conference on Learning Representations (ICLR)},
  2015.

\bibitem{lecun1998gradient}
Y.~LeCun, L.~Bottou, Y.~Bengio, and P.~Haffner.
\newblock Gradient-based learning applied to document recognition.
\newblock {\em Proceedings of the IEEE}, 86(11):2278--2324, 1998.

\bibitem{liu2017unsupervised}
M.-Y. Liu, T.~Breuel, and J.~Kautz.
\newblock Unsupervised image-to-image translation networks.
\newblock In {\em {Neural Information Processing Systems (NIPS)}}, 2017.

\bibitem{Liu:coGAN:NIPS16}
M.-Y. Liu and O.~Tuzel.
\newblock Coupled generative adversarial networks.
\newblock In {\em {Neural Information Processing Systems (NIPS)}}. 2016.

\bibitem{maaten2008visualizing}
L.~v.~d. Maaten and G.~Hinton.
\newblock Visualizing data using t-sne.
\newblock {\em Journal of Machine Learning Research}, 9(Nov):2579--2605, 2008.

\bibitem{mao2016multi}
X.~Mao, Q.~Li, H.~Xie, R.~Y. Lau, and Z.~Wang.
\newblock Multi-class generative adversarial networks with the l2 loss
  function.
\newblock {\em arXiv preprint arXiv:1611.04076}, 2016.

\bibitem{Mirza:cGAN:arXiv2014}
M.~Mirza and S.~Osindero.
\newblock Conditional generative adversarial nets.
\newblock {\em arXiv preprint arXiv:1411.1784}.

\bibitem{Moiseev:2013}
B.~Moiseev, A.~Konev, A.~Chigorin, and A.~Konushin.
\newblock Evaluation of traffic sign recognition methods trained on
  synthetically generated data.
\newblock In {\em International Conference on Advanced Concepts for Intelligent
  Vision Systems (ACIVS)}, 2013.

\bibitem{Morvant15}
E.~Morvant.
\newblock Domain adaptation of weighted majority votes via perturbed
  variation-based self-labeling.
\newblock {\em Pattern Recognition Letters}, 51:37--43, 2015.

\bibitem{netzer2011reading}
Y.~Netzer, T.~Wang, A.~Coates, A.~Bissacco, B.~Wu, and A.~Y. Ng.
\newblock Reading digits in natural images with unsupervised feature learning.
\newblock In {\em NIPS workshop on deep learning and unsupervised feature
  learning}, volume 2011, page~5, 2011.

\bibitem{pmlr-v70-odena17a}
A.~Odena, C.~Olah, and J.~Shlens.
\newblock Conditional image synthesis with auxiliary classifier {GAN}s.
\newblock In {\em Proceedings of the International Conference on Machine
  Learning (ICML)}, volume~70, 2017.

\bibitem{saito2017asymmetric}
K.~Saito, Y.~Ushiku, and T.~Harada.
\newblock Asymmetric tri-training for unsupervised domain adaptation.
\newblock In {\em International Conference on Machine Learning, (ICML)}, 2017.

\bibitem{Salimans:improvedGAN:NIPS16}
T.~Salimans, I.~J. Goodfellow, W.~Zaremba, V.~Cheung, A.~Radford, and X.~Chen.
\newblock Improved techniques for training gans.
\newblock In {\em {Neural Information Processing Systems (NIPS)}}, 2016.

\bibitem{sankaranarayanan2017generate}
S.~Sankaranarayanan, Y.~Balaji, C.~D. Castillo, and R.~Chellappa.
\newblock Generate to adapt: Aligning domains using generative adversarial
  networks.
\newblock {\em arXiv preprint arXiv:1704.01705}, 2017.

\bibitem{TRUDA-NIPS16_savarese}
O.~Sener, H.~O. Song, A.~Saxena, and S.~Savarese.
\newblock Learning transferrable representations for unsupervised domain
  adaptation.
\newblock In {\em Advances in Neural Information Processing Systems (NIPS)},
  pages 2110--2118. 2016.

\bibitem{Shrivastava:cvpr:17}
A.~Shrivastava, T.~Pfister, O.~Tuzel, J.~Susskind, W.~Wang, and R.~Webb.
\newblock Learning from simulated and unsupervised images through adversarial
  training.
\newblock In {\em Conference on Computer Vision and Pattern Recognition
  (CVPR)}, 2017.

\bibitem{be549cb97c19415788025989a4886d86}
J.~Stallkamp, M.~Schlipsing, J.~Salmen, and C.~Igel.
\newblock {\em The German traffic sign recognition benchmark: a multi-class
  classification competition}.
\newblock IEEE, 2011.

\bibitem{Sun:CORAL:AAAI16}
B.~Sun, J.~Feng, and K.~Saenko.
\newblock Return of frustratingly easy domain adaptation.
\newblock In {\em Conference of the Association for the Advancement of
  Artificial Intelligence (AAAI)}, 2016.

\bibitem{Taigman2016UnsupervisedCI}
Y.~Taigman, A.~Polyak, and L.~Wolf.
\newblock Unsupervised cross-domain image generation.
\newblock In {\em International Conference on Learning Representations (ICLR)},
  2017.

\bibitem{Tzeng_ICCV2015}
E.~Tzeng, J.~Hoffman, T.~Darrell, and K.~Saenko.
\newblock Simultaneous deep transfer across domains and tasks.
\newblock In {\em International Conference in Computer Vision (ICCV)}, 2015.

\bibitem{Hoffman:Adda:CVPR17}
E.~Tzeng, J.~Hoffman, T.~Darrell, and K.~Saenko.
\newblock Adversarial discriminative domain adaptation.
\newblock In {\em Computer Vision and Pattern Recognition (CVPR)}, 2017.

\bibitem{Wang:2004:IQA}
Z.~Wang, A.~C. Bovik, H.~R. Sheikh, and E.~P. Simoncelli.
\newblock Image quality assessment: From error visibility to structural
  similarity.
\newblock {\em Trans. Img. Proc.}, 13(4):600--612, Apr. 2004.

\bibitem{CycleGAN2017}
J.-Y. Zhu, T.~Park, P.~Isola, and A.~A. Efros.
\newblock Unpaired image-to-image translation using cycle-consistent
  adversarial networks.
\newblock {\em arXiv preprint arXiv:1703.10593}, 2017.

\end{thebibliography}
}

\end{document}